\documentclass[letterpaper, 10 pt, conference]{ieeeconf}  

\IEEEoverridecommandlockouts                              

\title{Cascaded Compositional Residual Learning \\ for Complex Interactive Behaviors}
\usepackage{mathrsfs}
\usepackage{amsmath}
\usepackage{amsfonts}
\usepackage[table,xcdraw,usenames,dvipsnames]{xcolor}
\usepackage{wrapfig}
\usepackage{soul}
\usepackage{longtable}
\usepackage{subcaption}
\usepackage{mwe}
\usepackage{multirow}

\newcommand{\irfan}[1]{\textcolor{red}{{}}}

\long\def\ignorethis#1{}

\newcommand{\E}[2]{\operatorname{\mathbb{E}}_{#1}\left[#2\right]}

\makeatletter
\let\oldlt\longtable
\let\endoldlt\endlongtable
\def\longtable{\@ifnextchar[\longtable@i \longtable@ii}
\def\longtable@i[#1]{\begin{figure}[t]
\onecolumn
\begin{minipage}{0.5\textwidth}
\oldlt[#1]
}
\def\longtable@ii{\begin{figure}[t]
\onecolumn
\begin{minipage}{0.5\textwidth}
\oldlt
}
\def\endlongtable{\endoldlt
\end{minipage}
\twocolumn
\end{figure}}
\makeatother

\usepackage{longtable}
\usepackage{booktabs}
\usepackage[final]{changes} 
\usepackage{hyperref}
\usepackage{url}
\usepackage{algorithm}
\usepackage{algpseudocode}

\let\labelindent\relax  
\usepackage{enumitem}
\author{K. Niranjan Kumar, Irfan Essa and Sehoon Ha
\thanks{This work was partly supported by the Cisco Research Corporation}
\thanks{K. Niranjan Kumar is with the Georgia Institute of Technology, Atlanta, GA 30332}%
\thanks{Sehoon Ha is with the Georgia Institute of Technology, Atlanta, GA 30332
USA, and with Robotics, Google, Mountain View, CA 94043 USA}%
\thanks{Irfan Essa is with the Georgia Institute of Technology, Atlanta, GA 30332
USA, and with Google, Mountain View, CA 94043 USA}%
}

\begin{document}

\maketitle
\thispagestyle{empty}
\pagestyle{empty}


\begin{abstract}
Real-world autonomous missions often require rich interaction with nearby objects, such as doors or switches, along with effective navigation. However, such complex behaviors are difficult to learn because they involve both high-level planning and low-level motor control. We present a novel framework, Cascaded Compositional Residual Learning (CCRL), which learns composite skills by recursively leveraging a library of previously learned control policies. Our framework learns multiplicative policy composition, task-specific residual actions, and synthetic goal information simultaneously while freezing the prerequisite policies. We further explicitly control the style of the motion by regularizing residual actions. We show that our framework learns joint-level control policies for a diverse set of motor skills ranging from basic locomotion to complex interactive navigation, including navigating around obstacles, pushing objects, crawling under a table, pushing a door open with its leg, and holding it open while walking through it. The proposed CCRL framework leads to policies with consistent styles and lower joint torques, which we successfully transfer to a real Unitree A1 robot without any additional fine-tuning. 
See videos at \href{https://www.kniranjankumar.com/ccrl/}{https://www.kniranjankumar.com/ccrl/}.

\end{abstract}

\section{INTRODUCTION}
Real-world autonomous missions often involve various levels of motor skills ranging from simple locomotion and manipulation to interactive navigation. For instance, a quadrupedal tasked to fetch an object may need to walk to a nearby door, open it with its end-effector, and navigate to the destination. Traditionally, many researchers ~\cite{sentis2006whole,ferrolho2022roloma,zimmermann2021go} have approached modeling such high-level behaviors by manually decomposing them into several low-level motor skills, such as locomotion, navigation, manipulation, and a high-level decision layer to modulate these skills. While effective, this model-based approach requires researchers to derive the explicit model of environmental interactions that are often complicated and cumbersome in such scenarios.

Learning-based approaches, such as deep reinforcement learning (deep RL)~\cite{arulkumaran2017deep,sutton2018reinforcement,kim2022human}, hold the promise of obtaining effective motor policies automatically from a simple description. However, learning complex behaviors is not straightforward due to many theoretical and practical challenges. For instance, motor creatures with many degrees of freedom, such as quadrupedal or bipedal robots, require a massive amount of simulation samples even for the simplest motor tasks due to their high-dimensional state and action spaces~\cite{tan2018sim,miki2022learning,kumar2021rma}. In addition, learning a high-level motor skill requires careful reward engineering, which is extremely time-consuming to tune. Researchers often adopt an explicit two-level hierarchical architecture to alleviate this problem~\cite{yang2020multi,li2020hrl4in}, but it only rearranges the order of the existing behaviors. For instance, it cannot combine locomotion and target-reaching into a successful door-opening skill.

\begin{figure}[t]
    \centering
    \includegraphics[width=0.48\textwidth]{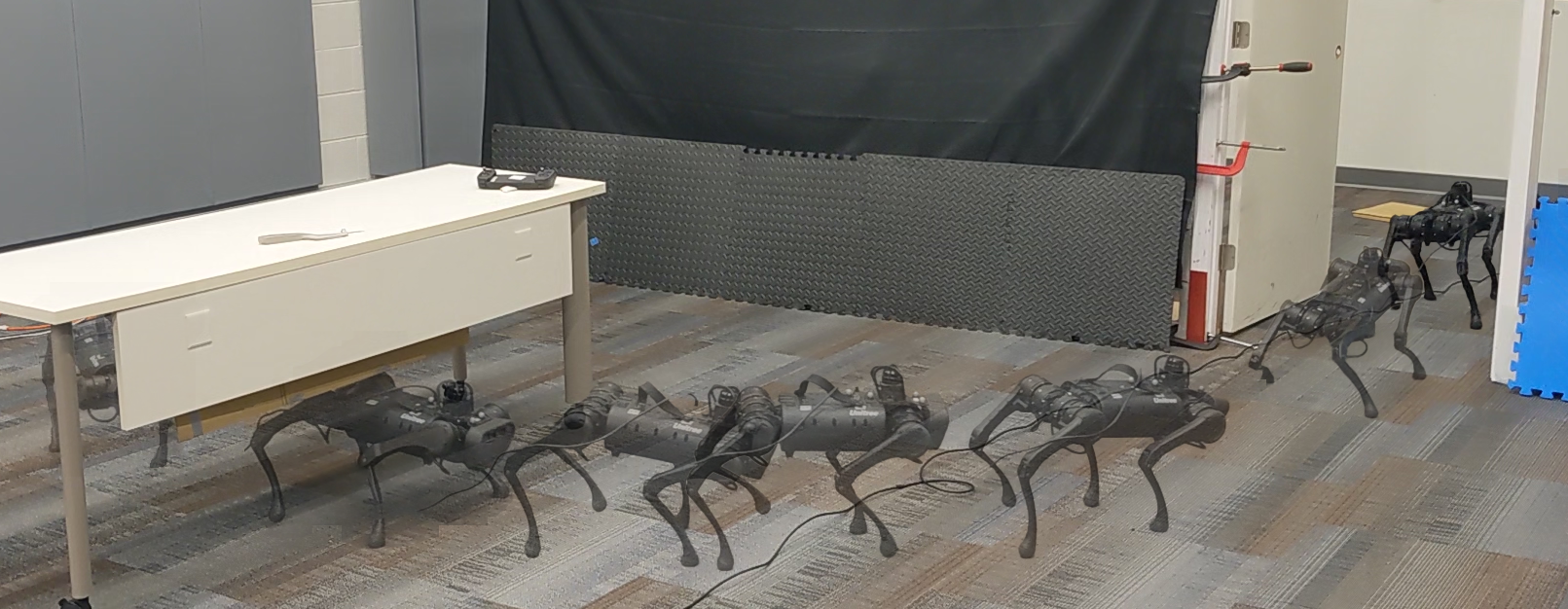}
    \caption{Our framework allows robots to learn complex environment interactions with a recursive hierarchy. The image shows a motion frame of the robot crawling under a table, reaching the door, pushing it open, and walking through it to reach a target location.}
    \vspace{-5pt}
    \label{fig:real_a1_door_open}
\end{figure}

We present a novel framework, Cascaded Compositional Residual Learning (CCRL), to learn a family of motor skills on a $12$ DOF quadruped robot to interactively navigate around an indoor scene. Inspired by residual learning~\cite{silver2018residual,johannink2019residual}, our key idea is to effectively train policies for a new challenging task via a cascaded multi-step process by recursively leveraging a set of pre-learned skills. We accomplish this by learning 1) a weighting network that composes pre-learned skills depending on the state of the agent in the environment, 2) a goal synthesis network that intelligently modulates the pre-learned skills, and c) a residual action network that learns task-specific perturbations to actions proposed by the pre-learned skills. Further, we introduce \emph{residual regularization} to control the trade-off between accomplishing a particular task efficiently vs. adhering to a combination of foundational skills to guarantee feasible motion on a real robot.

The main contributions of our work are as follows:
\begin{enumerate}
    \item A cascaded compositional residual learning (CCRL) framework to recursively train skills with increasing complexity by reusing skills learned in previous steps.
    \item A constrained residual learning objective that learns new skills while adhering to a style that is enforced by the set of pre-learned library of skills.
    \item Demonstration of interaction skills learned in simulation and successful transfer to a real Unitree A1 robot.
\end{enumerate}


\section{RELATED WORK}
\subsection{Robotic Locomotion}
Controlling legged robots has been a long-standing topic of research in the robotics community. Traditionally, roboticists ~\cite{apgar2018fast,hutter2016anymal,bledt2018cheetah,di2018dynamic,raibert1990trotting,kim2019highly}
have approached this problem with a combination of different frameworks and algorithms like trajectory optimization, model-predictive and whole-body control to demonstrate robust and agile locomotion. 
However, such manual controller design techniques often require in-depth prior knowledge about the robot dynamics, which restricts its applicability to new problems. 
On the other hand, reinforcement learning~\cite{sutton2018reinforcement} offers an automated controller design process by optimizing a policy for a reward function that measures performance on a given task. 
A body of work~\cite{fu2021minimizing, hafner2020towards,kumar2021rma,miki2022learning,tan2018sim,fu2022coupling,hwangbo2019learning,rudin2022learning} in RL has demonstrated effective learning of locomotion policies. 
However, these learned policies often show degraded performance on hardware due to the difference between the simulation and the real world, referred to as the \emph{sim-to-real} gap. 
Several techniques have been proposed to bridge the sim-to-real gap, such as domain randomization \cite{tan2018sim,miki2022learning}, learning actuator dynamics \cite{hwangbo2019learning}, online adaptation \cite{yu2020learning,kumar2021rma}, real-world learning \cite{ha2020learning,smith2022legged}. 
Our work also leverages existing domain randomization techniques to deploy learned interactive behaviors to the real world.

\subsection{Interactive Navigation}
Navigation is a fundamental skill for autonomous robot missions. Several navigation problems have been proposed over the years~\cite{anderson2018evaluation,li2020hrl4in}, such as: PointNav - navigating to a point in a map; ObjectNav - navigating to a selected object category in the scene; InteractiveNav - Navigating to a point in a scene that requires interaction with objects and furniture in the scene.
While researchers have made significant progress in PointNav~\cite{wijmans2019dd} and ObjectNav~\cite{sorokin2021learning,chaplot2020object,batra2020objectnav}, InteractiveNav~\cite{li2020hrl4in,xia2020interactive,zeng2021pushing,konidaris2011autonomous} is still a challenging problem due to the difficulty in learning interaction dynamics between a robot and its environment.
Li et al. ~\cite{li2020hrl4in} learned a navigation policy on a mobile robot that can open a door to reach its target location. 
Konidaris et al.~\cite{konidaris2011autonomous} took an alternate approach where the robot was given access to a set of hard-coded controllers and had to learn when to use each skill.
However, these approaches focus on just wheeled mobile robots, abstracted to be a simple cylinder with an attached manipulator, significantly limiting the space of possible interactions.
Sunwoo et al.~\cite{kim2022human} demonstrated interaction skills on a quadrupedal robot via manual motion-based control.
In contrast, our work focuses on learning autonomous policies to directly control joint motors of a 12-DOF quadruped robot, to perform complex dynamic interactions such as door opening, object pushing and crawling under a table. To the best of our knowledge, our work is the first of it's kind, demonstrating end-to-end neural network policies, that can solve InteractiveNav on a high DOF legged robot, with a high success rate.


Another related problem is interactive search, where a robot has to \textit{search} for an object in an environment. 
Most of the work in this area~\cite{kumar2022graph,Danielczuk2020XRayMS,huang2020mechanical} focuses on searching for an object in a cluttered shelf or table. 
In a more generalized version of interactive search, a robot moves around a cluttered indoor scene searching for an object. 
While we do not tackle interactive search, our method complements existing techniques by enabling a robot \textit{interactively} navigate to the object, once its location has been estimated.

\subsection{Hierarchical Reinforcement Learning}
Learning RL policies for complex tasks requires extensive reward engineering and hyperparameter tuning. A common technique to deal with such complexity is to decompose the policy into multiple sub-tasks.
In hierarchical RL (HRL)~\cite{bacon2017option,sutton1999between}, higher level policies control and instruct low level policies like they were primitive actions. 
While an in-depth discussion about hierarchical RL is outside the scope of this paper, we point interested readers to refer Pateria et al.~\cite{pateria2021hierarchical}, and restrict our discussion to related work in locomotion and navigation. 
A common approach to navigation is to use low-level locomotion primitives in combination with a high-level primitive selection or goal proposal network to perform navigation~\cite{li2020learning,jain2019hierarchical,heess2016learning,peng2017deeploco}. Alternatively, Yang et al.~\cite{yang2020multi} proposed a gating network that combines the parameters of a collection of neural network primitives, to produce composite navigation policy.
Similar to these works, we are interested in designing a hierarchy to solve a complex task, but our low-level policies are themselves complex (like opening a door) and not easily breakable into simpler policies. 
Therefore, a common ``flat'' hierarchy will not be sufficient to learn more challenging motor skills.
We propose a multi-level cascaded hierarchy to build progressively complex policies starting out from a small set of learned skills. 
We build this hierarchy on the fly, by learning residuals~\cite{silver2018residual,johannink2019residual} that can perturb the output of an existing library of policies, to generate novel behaviors. 



\section{Cascaded Compositional Residual Learning}

This section describes our cascaded compositional residual learning (CCRL) framework that learns complex motor behaviors, by recursively obtaining control policies and reusing them to solve increasingly difficult tasks. 
We will begin the section by explaining relevant background, followed by problem formulation, skill reusing mechanisms and learning algorithms.

\subsection{Background: Markov Decision Process}
\label{sec:mdp}
Robot learning can be modeled as a Markov Decision Process (MDP) defined by the 5-tuple $ \mathcal{M} : \langle \mathcal{S},\mathcal{A},\mathcal{T}, r,\gamma \rangle$, where $\mathcal{S}$, $\mathcal{A}$ and $r$ are the state space, the action space, and the reward function, respectively. $\mathcal{T}$ is a transition function that determines the next state given the current state and action. $\gamma \in (0,1)$ is the discount factor. Our goal then, is to find a policy $\pi: \mathcal{S} \rightarrow \mathcal{A}$ that maximizes a cumulative return $J(\pi) = \E{\tau\sim \rho_\pi}{\sum_{t = 0}^{T} r(s_t, a_t)}$, where $\tau$ is the distribution of the generated trajectories.

However, solving the given MDPs for high-level motor tasks, such as interactive navigation or full-body manipulation, is not straightforward. 
First, the given behavior involves multiple sub-tasks, such as walking or door opening, which require the manual design of very specific reward functions. 
In addition, the long-horizon nature of the problem makes it extremely sensitive to the choice of hyper-parameters. 
Further, we want to maintain smooth and stable motion styles to make them consistently feasible on actual hardware. 
Therefore, it is almost impossible to obtain such good motion controllers for these challenging motor tasks via simple reward engineering and hyperparameter tuning.



\subsection{Background: Residual Learning}
Residual learning~\cite{johannink2019residual,silver2018residual} offers a promising alternative to the traditional training-from-scratch RL framework by enabling the transfer of a skill learned on one task to another related but more difficult task. Consider a policy $\pi_{0}$ to be a solution for the base task. We can freeze the parameters of $\pi_{0}$ and learn a residual policy $\pi_k$ to perturb the actions of $\pi_{0}$, modify learned behaviors and accomplish a different goal. The final policy is then obtained by adding the outputs of these two policies:
\begin{equation*}
    \pi(s) = \pi_0(s) + \pi_k(s)
\end{equation*}


\subsection{Problem Definition: Multi-skill Learning}
\begin{figure}
\vspace{0.1in}
\centering
\includegraphics[width=0.85\linewidth]{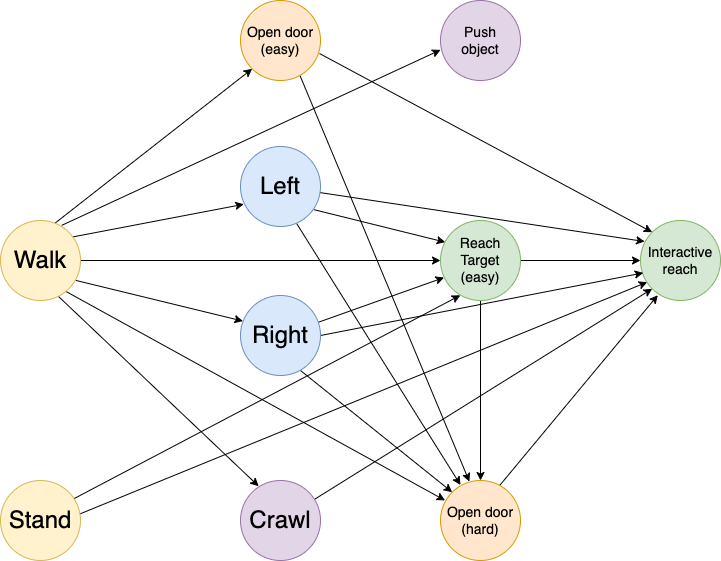}
\caption{Directed graph representing the relationship between skills. Each skill is built as a compositional policy over its parent policies and a learned residual specific to that skill.}
\label{fig:skill_relationships}
\end{figure}
In our formulation, we consider a set of $n$ motor tasks described as MDPs $\mathrm{\Omega}=\{\mathcal{M}_0,\mathcal{M}_1,\mathcal{M}_2, \dots,\mathcal{M}_n \}$, where each $\mathcal{M}_k: \langle \mathcal{S}_k,\mathcal{A},\mathcal{T}, r_k,\gamma \rangle \quad \forall \quad 0<k<n$, models the MDP for a learnable skill. $\mathcal{S}_k,\mathcal{A},\mathcal{T}, r_k$ and $\gamma$ represent the state space, action space, transition function, reward function and discount factor respectively. 
Note that we allow unique state spaces for the problems, which indicates that some high-level behaviors require additional inputs for describing the task (e.g., the target location) or environmental states (e.g., the door hinge angle).  
On the other hand, we assume a single robot, which leads to unified actions and transition functions.

We then want to learn a collection of control policies  $\mathcal{C}=\{\pi_{1}, \pi_{2}, \pi_{3}, \dots, \pi_n\}$. A na\"ive approach is to independently solve each corresponding MDP. 
But in practice, it is often not feasible due to the challenges outlined in Section~\ref{sec:mdp}. Our key insight is to ``recursively'' learn policies for challenging MDPs on top of the related ``prerequisite'' or ``parent'' skills instead of learning from scratch. 
To this end, we assume a Skill Decomposition Graph $\mathcal{G}$ by defining a set of prerequisite skills, where its vertices are the corresponding MDPs, and its edges represent the dependencies. 
For instance, Figure~\ref{fig:skill_relationships}, shows the dependencies of \textit{Reach Target (Easy)} on \textit{Walk}, \textit{Stand}, \textit{Turn Left} and \textit{Turn Right}. 

We assume the skill decomposition graph $\mathcal{G}$ to be intuitive to define (e.g., navigation requires straight walking and turning) but could also be automatically estimated by searching the space of all possible relations and retaining those that yield the best performance.
Further, our framework is robust to the redundancy in prerequisite skills in the sense that it learns with subset of them or ignore irrelevant prerequisite skills (e.g., if the task does not require the robot to crawl, the crawling skill will be ignored). Please refer to Section~\ref{sec:analysis}.

\subsection{Cascaded Compositional Residual Learning}





\begin{figure*}
    \centering
    \vspace{0.1in}
    \includegraphics[width=0.9\textwidth]{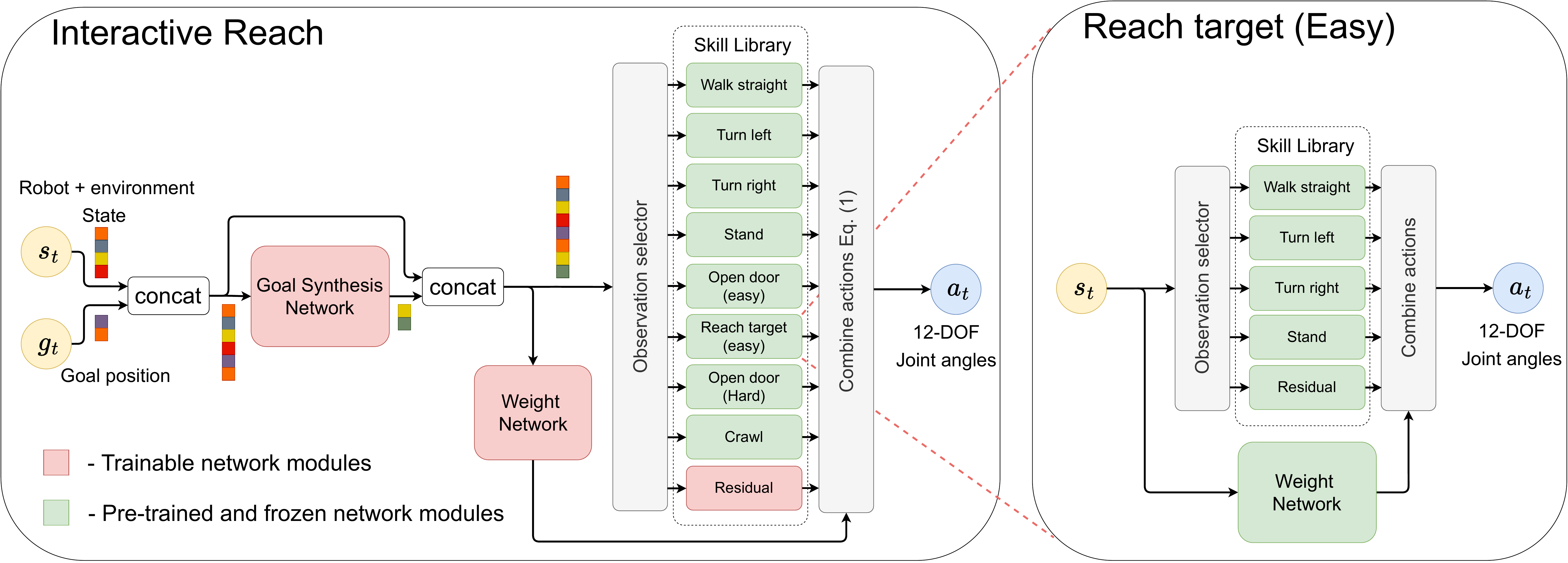}
    \caption{Overview of our Cascaded Compositional Residual Learning, which consists of three learnable components: a synthetic goal network, a weight network, and a residual action network. The example illustrates the architecture for \textit{Interactive Reach}, which uses \textit{Reach target (Easy)} on top of other primitive skills. 
    }
    \label{fig:cascaded_architecture}
\end{figure*}

A traditional flat architecture of HRL would train each of these policies from scratch, to solve the corresponding MDP. But due to challenges outlined in the previous section, it is not practically feasible. The key insight of our approach is that, instead of learning new skills from scratch, we build them on top of pre-learned skills, i.e. $\pi_k$ could be a composite policy over the prerequisite policies $\mathcal{C}_k$, where $\mathcal{C}_k \subset \mathcal{C}$. Mathematically,
\begin{equation*}
    \pi_k = \mathcal{F}_k(\mathcal{C}_k\cup\{{\pi_k^r}\})
\end{equation*}
where $\mathcal{F}_k$ is a merging function that combines the outputs of all $\pi_j \quad \forall \quad \pi_{j}\in \mathcal{C}_k$ and $\pi_k^r$ is a residual neural network policy. To this end, our framework (Figure~\ref{fig:cascaded_architecture}) has three trainable networks, (1) a residual action network,  (2) a weight network, and (3) a synthetic goal network, while freezing the prerequisite policies in $\mathcal{C}_k$.

\noindent\textbf{Residual Network.} The residual policy $\pi_k^r$ generates a \textit{residual action} that perturbs the actions generated by the learned policies to generate novel behavior that is absent in the set of pre-learned skills. 
We simply treat this residual network as an additional policy: the only difference being that the parameters of $\pi_k^r$ are learnable during policy training.

\noindent\textbf{Weight Network.}
While prior work in residual learning combines two policies by just adding them together, this approach has been observed to be sub-optimal by Peng et al.~\cite{peng2019mcp} when combining a large set of skills. They proposed Multiplicative Compositional Policy (MCP) that combines a set of stochastic policies while ensuring that multiple policies simultaneously work together and provide more flexibility in the final composite policy. The MCP objective combines multiple skills by using the following equation:
\begin{equation}
\pi(a \mid s, g)=\frac{1}{Z(s, g)} \prod_{i=0}^{k} \pi_{i}(a \mid s, g)^{w_{i}(s, g)}, 
\end{equation}
where $w_i(s,g)$ are learned weighting functions and $Z(s,g)$ is a normalizing factor to ensure that the weights are between $0$ and $1$.

\noindent \textbf{Goal Synthesis Network.} This network creates intermediate goals for individual skills. For example, the \textit{Reach Target (Easy)} skill needs a goal location around the robot as input. But to use this skill as part of \textit{Door open (Hard)}, we need to generate a collection of intermediate targets as input the the policy. These targets are generated by the Synthetic Goals Network.


The architecture is illustrated in Figure~\ref{fig:cascaded_architecture}. Our work provides framework that builds novel complex skills, by leveraging a library of pre-learned composite skills, while ensuring controllability over the style of the composite. We use our framework to learn a wide range of interactive skills on a quadruped robot while implicitly guiding the \textit{style} of the policy by controlling the weight given to the residual actions. With this framework we train a handful of basic policies from scratch, using RL, and use these policies to progressively build increasingly complex composite skills. 

\subsection{Auxiliary Loss}
\label{residual_penalty}
\begin{figure*}[ht]
    \vspace{0.1in}
    \begin{subfigure}[h]{0.19\linewidth}
    \includegraphics[width=0.99\textwidth]{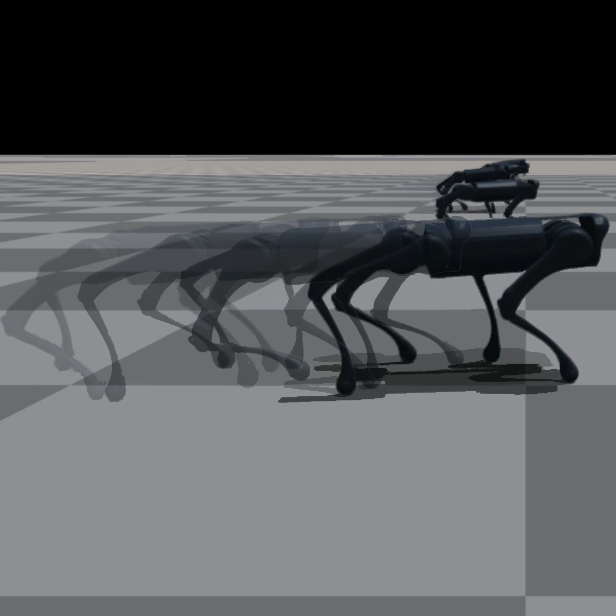}
    \caption{Straight walking}
    \label{fig:snapshot_walk}
    \end{subfigure}
    \begin{subfigure}[h]{0.19\linewidth}
    \includegraphics[width=0.99\textwidth]{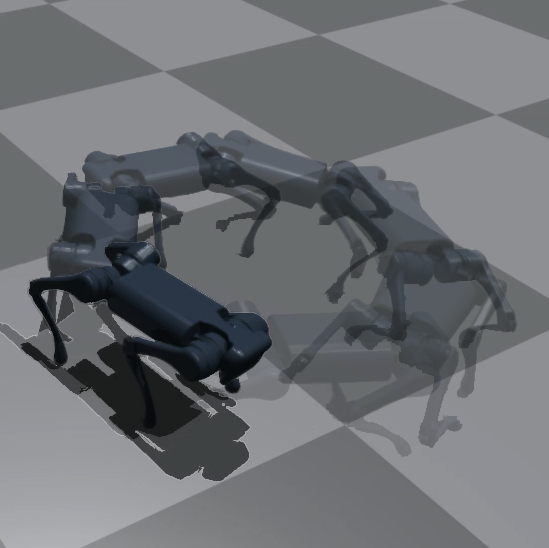}
    \caption{Turning}
    \label{fig:snapshot_turn}
    \end{subfigure}
    \begin{subfigure}[h]{0.19\linewidth}
    \includegraphics[width=0.99\textwidth]{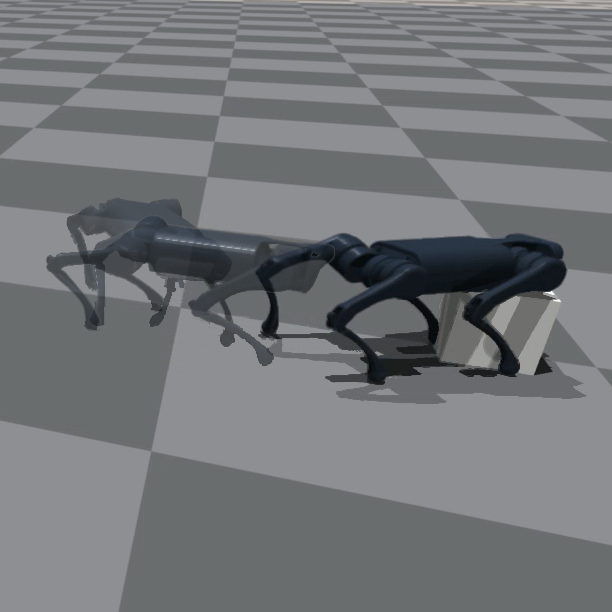}
    \caption{Reach Target (Easy)}
    \label{fig:snapshot_reach_target}
    \end{subfigure}
    \begin{subfigure}[h]{0.19\linewidth}
    \includegraphics[width=0.99\textwidth]{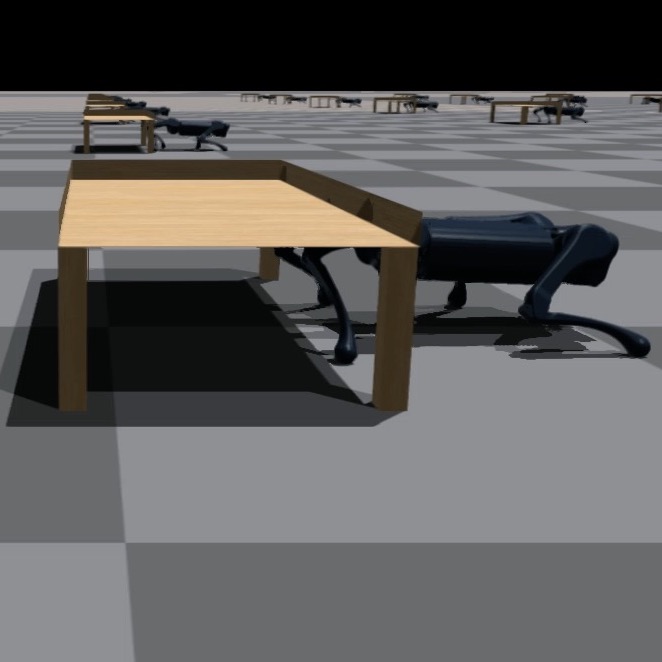}
    \caption{Crawling}
    \label{fig:snapshot_crawl}
    \end{subfigure}
    \begin{subfigure}[h]{0.19\linewidth}
    \includegraphics[width=0.99\textwidth]{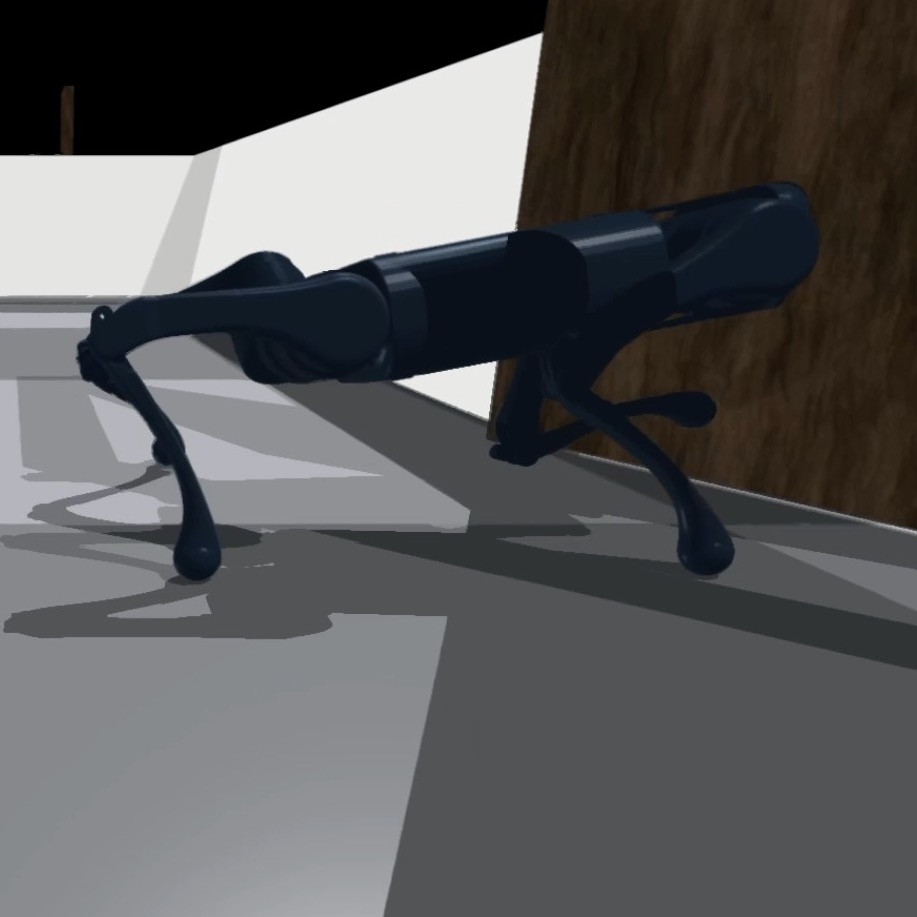}
    \caption{Door Open (Easy)}
    \label{fig:snapshot_door_easy}
    \end{subfigure}
    
    \begin{subfigure}{0.24\linewidth}
    \includegraphics[width=0.99\textwidth]{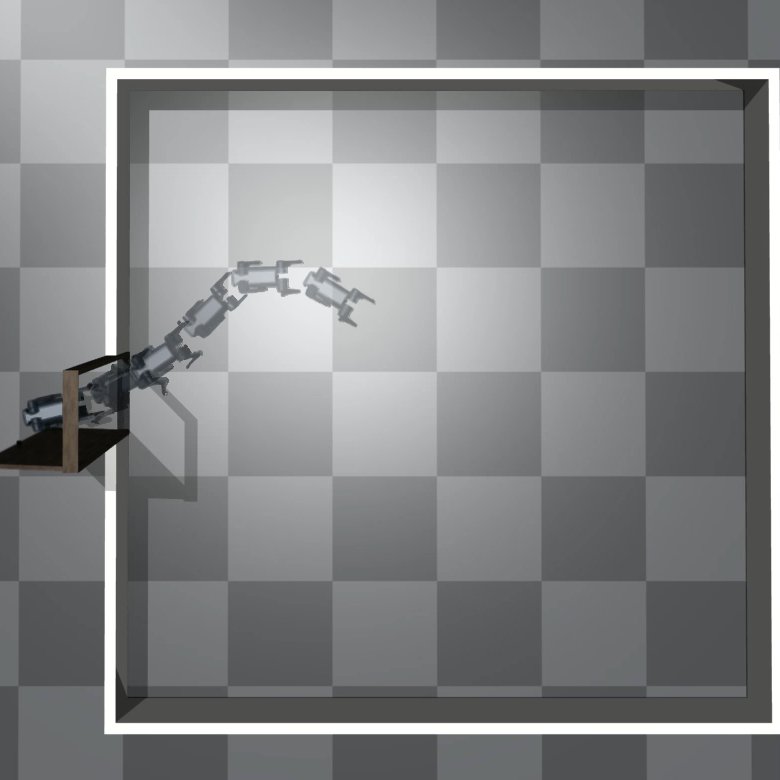}
    \caption{Door Open (Hard)}
    \label{fig:snapshot_door_hard}
    \end{subfigure}
    \begin{subfigure}{0.24\linewidth}
    \includegraphics[width=0.99\textwidth]{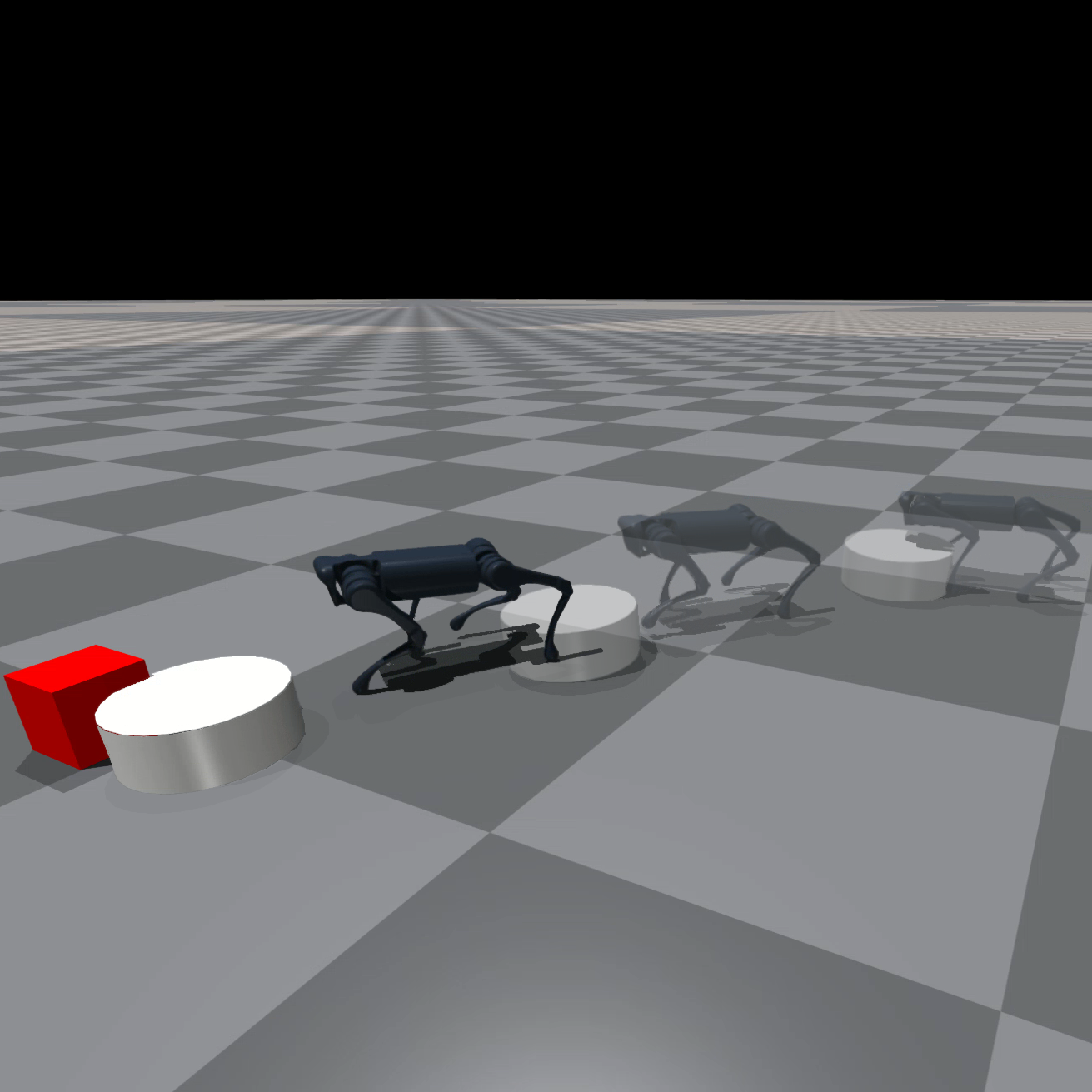}
    \caption{Push Object to Target}
    \label{fig:snapshot_push}
    \end{subfigure}
    \begin{subfigure}{0.50\linewidth}
    \includegraphics[width=0.95\textwidth]{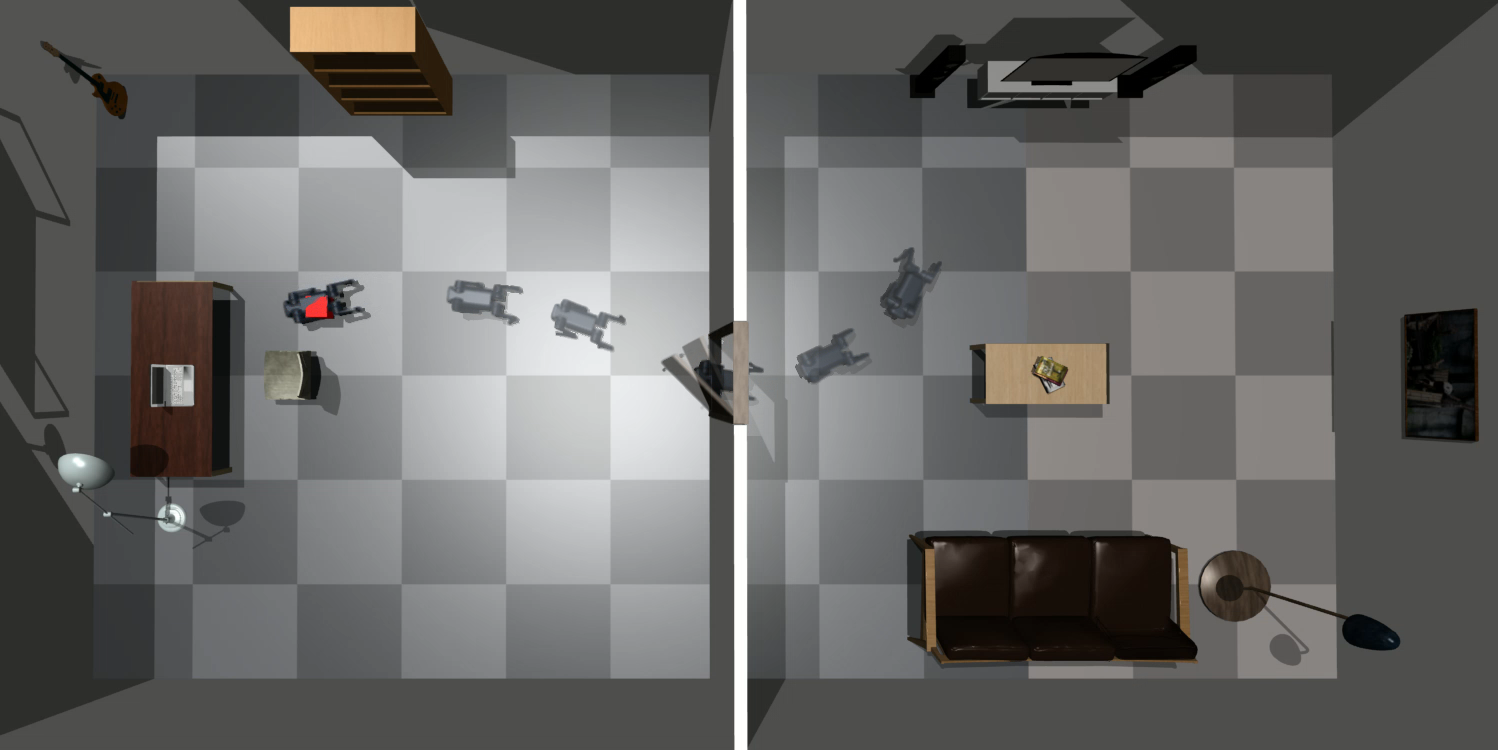}
    \caption{Interactive Reach}
    \label{fig:interactive_reach}
    \end{subfigure}
    \caption{Overview of the tasks we train our robot to accomplish.}
    \label{fig:environment_snapshots}
\end{figure*}
In the previous section, we discussed learning novel skills by effectively leveraging a library of existing skills and a trainable residual. However, an unconstrained residual could dominate the output of the policy and give rise to unstable behaviors, which eventually lead to sub-optimal behaviors and unsuccessful sim-to-real transfer. Therefore, we introduce constraints into our policy optimization framework to control the extent to which residuals influence the policy. We modify the traditional RL objective with regularization terms to penalize the magnitude and weight given to the residual. 

\begin{equation*}
    L^{rw}_t = \mathbb{E}_t\left[{\mid w_{res}(s_t, g_t) \mid}_1\right]
\end{equation*}
\begin{equation*}
    L^{rm}_t = {\mid \mathbb{E}_t\left[\pi_{res}(a_t \mid s_t, g_t)\right] \mid}_1
\end{equation*}

$L^{rm}$ and $L^{rw}$ are penalties on the magnitude of the residual actions and the weights assigned to them respectively.

In our work, we use PPO (Proximal Policy Optimization)~\cite{schulman2017proximal} as the RL algorithm of choice. Consider the standard PPO (Proximal Policy Optimization)~\cite{schulman2017proximal} objective below:
\begin{equation*}
    L_t^{PPO}(\theta)=\hat{\mathbb{E}}_t\left[L_t^{C L I P}(\theta)-c_1 L_t^{V F}(\theta)+c_2 S\left[\pi_\theta\right]\left(s_t\right)\right]
\end{equation*}
where $L_t^{C L I P}(\theta)$ is the clipped policy gradient objective, $L_t^{V F}(\theta)$ is the value function error and  $S\left[\pi_\theta\right]\left(s_t\right)$ is the entropy term. $c_1, c_2$ are weights to control the influence of each term over the total loss. 

We modify this loss to include regularization terms defined above: 
\begin{equation}
    \label{eq:ppo_updated}
    L_t(\theta) = L_t^{PPO}(\theta) + c_3 L^{rw}_t + c_4 L^{rm}_t.
\end{equation}
We set coefficients $c_1$, $c_2$ to be $1.0$ and $0.01$ respectively. $c_3$, and $c_4$ are task dependent coefficients that are tuned to maximize the policy performance. 
Once we train a skill, we save the residual, weight, and goal synthesis networks and reuse them to train other skills. The critic is trained from scratch for every new skill.
\section{EXPERIMENTS}

In this section, we design experiments to test the effectiveness of the proposed method, cascaded compositional residual learning (CCRL), by learning skills involving different levels of difficulty and comparing the performance with traditional RL baselines. Next, we also present simulation analyis to investigate the robustness of the proposed method and the importance of the auxiliary loss. Finally, we examine the learned skill policies on a real Unitree A1 robot and demonstrate robust sim-to-real transfer. 

\subsection{Problem Formulation} \label{sec:tasks}

In our work, we train a collection of policies, that help an agent interactively navigate an indoor scene. Each skill has its own underlying MDP, but we provide a rough definition of the ingredients of the MDPs as follows. 

\noindent\textbf{States.} A 60-dimensional state space consists of the base linear velocity, base angular velocity, base yaw angle, base position, gravity vector projected on the base, joint angles, joint velocities, and object locations relative to the robot.

\noindent\textbf{Actions.} The action space is the desired target joint angles of the robot ($12$ for A1), which are fed into Proportional Derivative (PD) controllers.

\noindent\textbf{Rewards.} We use a collection of terms to account for different aspects of the robot motion. Let us denote the velocity as $\mathbf{v}$, the angular velocity as $\boldsymbol{\omega}$, the joint angles as $\mathbf{q}$, joint velocities as $\dot{\mathbf{q}}$, joint torques as $\boldsymbol{\tau}$, number of robot parts (excluding feet) in contact with the environment as $n_{contact}$, action taken at a given step as $a_t$, the angle made by the hinge of the door joint as $q_{door}$, the position of the target relative to the robot as $\mathbf{x}_{target}$, the position of the target relative to the object being pushed as $\mathbf{x}_{t2o}$ and $\Delta t$ as the simulation time-step. The reward at time $t$ is defined as the weighted sum of the following quantities:
\begin{enumerate}[leftmargin=0.9cm,label=(R\arabic*)]
    \item Linear velocity tracking:  $exp(-(\mathbf{v}_{target}-\mathbf{v})^2/\sigma_1)$
    \item Angular velocity tracking:  $exp(-(\omega_{z_{target}}-\omega_z)^2/\sigma_2)$
    \item Pitch and roll penalty:      $\omega_x^2+\omega_y^2$
    \item Joint acceleration penalty: $\sum(\frac{\dot{\mathbf{q}}_t-\dot{\mathbf{q}}_{t-1}}{\Delta t})^2$           
    \item Collision penalty: $n_{contact}$   
    \item Action change penalty : $\sum(\mathbf{a}_t-\mathbf{a}_{t-1})^2$   
    \item Torque penalty : $\sum \boldsymbol{\tau}^2$            
    \item Door angle: $q_{door}$                 
    \item Distance to target: $exp(-\left\|\mathbf{x}_{target} \right\|/\sigma_3 )$ 
    \item Object-target distance: $exp(-\left\|\mathbf{x}_{t2o} \right\|/\sigma_3 )$ 
\end{enumerate}
$\sigma_1$, $\sigma_2$, $\sigma_3$ are scaling factors that we set to 0.25, 0.25 and 2.0 respectively.


\noindent\textbf{Tasks.}
We define the following tasks, each aimed at developing a skill that is useful while navigating or exploring an indoor cluttered scene (Figure~\ref{fig:environment_snapshots}). Their relationship is defined by the skill graph $\mathcal{G}$ (Figure~\ref{fig:skill_relationships}).
\begin{enumerate}
    \item \textbf{Walk straight}: Walk forward in a straight line. 
    \item \textbf{Turn left}: Turn to the left. The episode is successful if the robot circle backs to its initial yaw.
    \item \textbf{Turn right}: Turn to the right. The episode is successful if the robot circle backs to its initial yaw.
    \item \textbf{Stand}: Stand in place when subjected to external disturbances. 
    \item \textbf{Reach target (Easy)}: Reach a target placed anywhere within a circle of radius $3m$ from the center of the robot. The episode is successful if the robot's center is within $0.5m$ from the goal (R9).
    \item \textbf{Open door (Easy)}: Push door open with foot and hold it open while walking through it. The episode is successful if the robot crosses the door (R8).
    \item \textbf{Open door (Hard)}: Reach the door from an arbitrary starting point in a room, push it open with a foot and hold it open while walking through it. The episode is successful if the robot crosses the door (R8).
    \item \textbf{Push object}: Push a cylindrical puck to a target location. The episode is successful if the puck is within $0.1m$ from the desired target location (R10).
    \item \textbf{Crawling}: Crawl under a slab that gradually decreases in height, while avoiding collisions. The episode is successful if the robot crawls out from under the slab.
    \item \textbf{Interactive Reach}: Reach a target object placed anywhere in a two-room house with furniture and a door, while avoiding collisions. Robot is successful if its body center within $0.5m$ from the center of the target (R9).
    
\end{enumerate}
All the tasks are trained with rewards R1-R7 in addition to the ones mentioned above. 

\subsection{Simulation Setup}
We use Isaac Gym~\cite{makoviychuk2021isaac} to simulate our interactive environment and train our policies. We run $4096$ environments in parallel, on a single NVIDIA Titan X GPU. During training, we randomize surface friction between the robot and the ground by randomly sampling from the range-$\left[0.5,1.25\right]$. We implement all our policies as fully connected neural network layers, two layers of $256$ to $512$ neurons depending on the task difficulties, with exponential linear units (\textit{Elu}). We train policies with our updated PPO objective (Equation ~\ref{eq:ppo_updated}). 

\begin{figure}[tp]
\vspace{0.1in}
    \centering
    \includegraphics[width=0.92\linewidth]{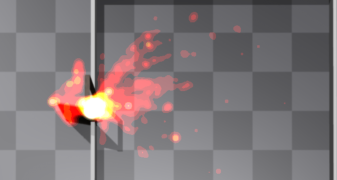}
    \caption{Heatmap showing the weight magnitude on the \textit{Open door (Easy)} skill. Notice the increased activation of the skill when the robot is close to the door.}
    \label{fig:skill_heatmap}
\end{figure}

\begin{table}[t]
\centering
\begin{tabular}{|c|cccc|}
\hline
                                 & \multicolumn{4}{c|}{Success Rate}                                                                                                                                                                                                            \\ \cline{2-5} 
                                 & \multicolumn{1}{c|}{\begin{tabular}[c]{@{}c@{}}Reach Target\\ (Easy)\end{tabular}} & \multicolumn{1}{c|}{\begin{tabular}[c]{@{}c@{}}Open Door \\ (Hard)\end{tabular}} & \multicolumn{1}{c|}{\begin{tabular}[c|]{@{}c@{}}Push \\ Object\end{tabular}} & \begin{tabular}[c]{@{}c@{}}Interactive\\ Reach\end{tabular} \\ \hline
CCRL                             & \multicolumn{1}{c|}{0.90}                                                          & \multicolumn{1}{c|}{\textbf{0.98}}                                               & \multicolumn{1}{c|}{\textbf{0.89}}    &    \textbf{0.72}                                                \\ \hline
Vanilla                          & \multicolumn{1}{c|}{\textbf{0.95}}                                                 & \multicolumn{1}{c|}{0}                                                           & \multicolumn{1}{c|}{0.04}              &       0.46                                            \\ \hline
Curriculum                       & \multicolumn{1}{c|}{0.92}                                                          & \multicolumn{1}{c|}{\textbf{0.98}}                                               & \multicolumn{1}{c|}{0.52}               &         0.48                                         \\ \hline
\multicolumn{1}{|l|}{Big Policy} & \multicolumn{1}{c|}{0.92}                                                          & \multicolumn{1}{c|}{0}                                                           & \multicolumn{1}{c|}{0.01}               &       0.22                                          \\ \hline
\end{tabular}
\caption{Performance comparison of our method against three baselines \emph{Vanilla}, \emph{Curriculum}, and \emph{Big Policy}. Our method achieves significantly better success rates on harder tasks, such as \textit{Push Object} or \textit{Interactive Reach}.}
\label{tab:success_rate}
\end{table}

\subsection{Simulation Results}
Our framework enables us to train complex long horizon interactive behaviors grounded in previously learned parent skills. Our policy was able to learn all the tasks described in Section~\ref{sec:tasks} and Figure~\ref{fig:environment_snapshots}. The success rates of the learned policies are near $90$\%, except for the hardest task of \textit{Interactive Reach} that shows a $72$\% success rate. We plot the heatmap of the residual weights on the \textit{Open Door (Easy)} skill with respect to its position in Figure~\ref{fig:skill_heatmap}. This figure demonstrates the importance of the weight network in learning a new skill.

\noindent\textbf{Performance Comparison.} To highlight the importance of our cascaded residual framework, we compare our method with three baselines: 
\begin{enumerate}
    \item \textbf{Vanilla:} The policy is trained from scratch to solve the given problem.
    \item \textbf{Curriculum:} A single policy is trained using a manually curriculum that gradually increases the complexity of the environment.
    \item \textbf{Big Policy}: To show that our improved performance is not a byproduct of architecture size, we train a policy that uses the same neural network architecture as our proposed policy, but is trained from scratch.
\end{enumerate}
We compare the success rates on multiple tasks in Table~\ref{tab:success_rate}. Our approach, CCRL, typically outperforms all the other three baselines, Vanilla, Curriculum, and Big Policy, on challenging tasks. For the \textit{Door Open (Hard)}, only CCRL and Curriculum achieve good success rates ($98$\%) while Vanilla and Big policy fail. In our experience, it is almost impossible to learn an effective door-opening skill without a proper curriculum, such as learning to walk and open a door. However, for the most complicated task of Interactive Reach, CCRL shows the best success rate of $72$\% while all three baselines only show less than $48$\% success rates. This is because Interactive Reach requires careful coordination of all multiple different skills, navigation, crawling, obstacle avoidance, and door manipulation, which shows the importance of cascaded skill learning.

\begin{figure}[tp]
\vspace{0.1in}
    \centering
    \includegraphics[width=0.99\linewidth]{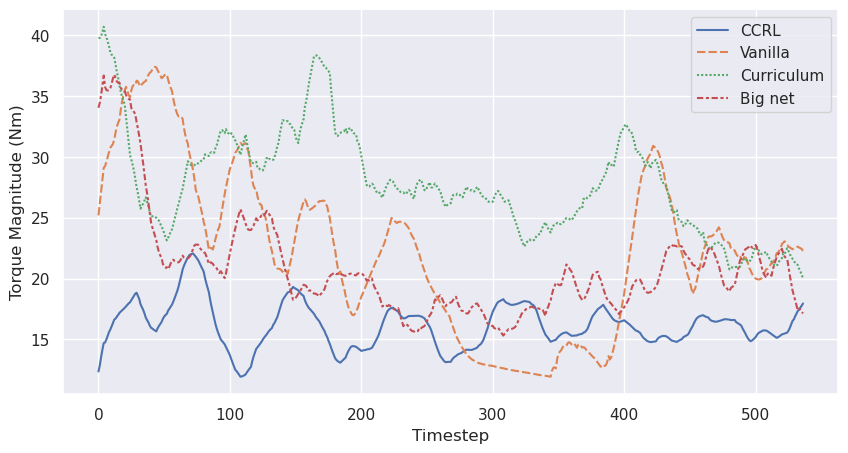}
    \caption{Torque curves of our approach compared to baselines. Although all techniques have a success rate of around $90-95\%$, our approach has a much lower net torque magnitude and trajectories better suited for sim-to-real transfer.}
    \label{fig:torques}
\end{figure}

\noindent\textbf{Control over Quality.} Further, because all our skills are grounded to preliminary skills, such as walking and standing, through residual regularization, we notice that the quality of motion is also significantly better. This feature gives us an explicit way to control the style of new policies. For instance, in the case of \textit{Reach Target (easy)}, the robot might jump toward the target or wiggle its legs to move gradually toward the target. While these behaviors might be successful in simulation, they do not transfer well to the real world as they are overfitted to the simulation dynamics and parameters. For illustration, we compare the torque trajectories in Figure~\ref{fig:torques}. Notice that our method uses a significantly lower average torque when compared to the baselines, which results in better motion quality. Please also refer to the supplemental video for qualitative comparison.

\begin{figure}[t]
\vspace{0.1in}
    \centering
    \includegraphics[width=0.98\linewidth]{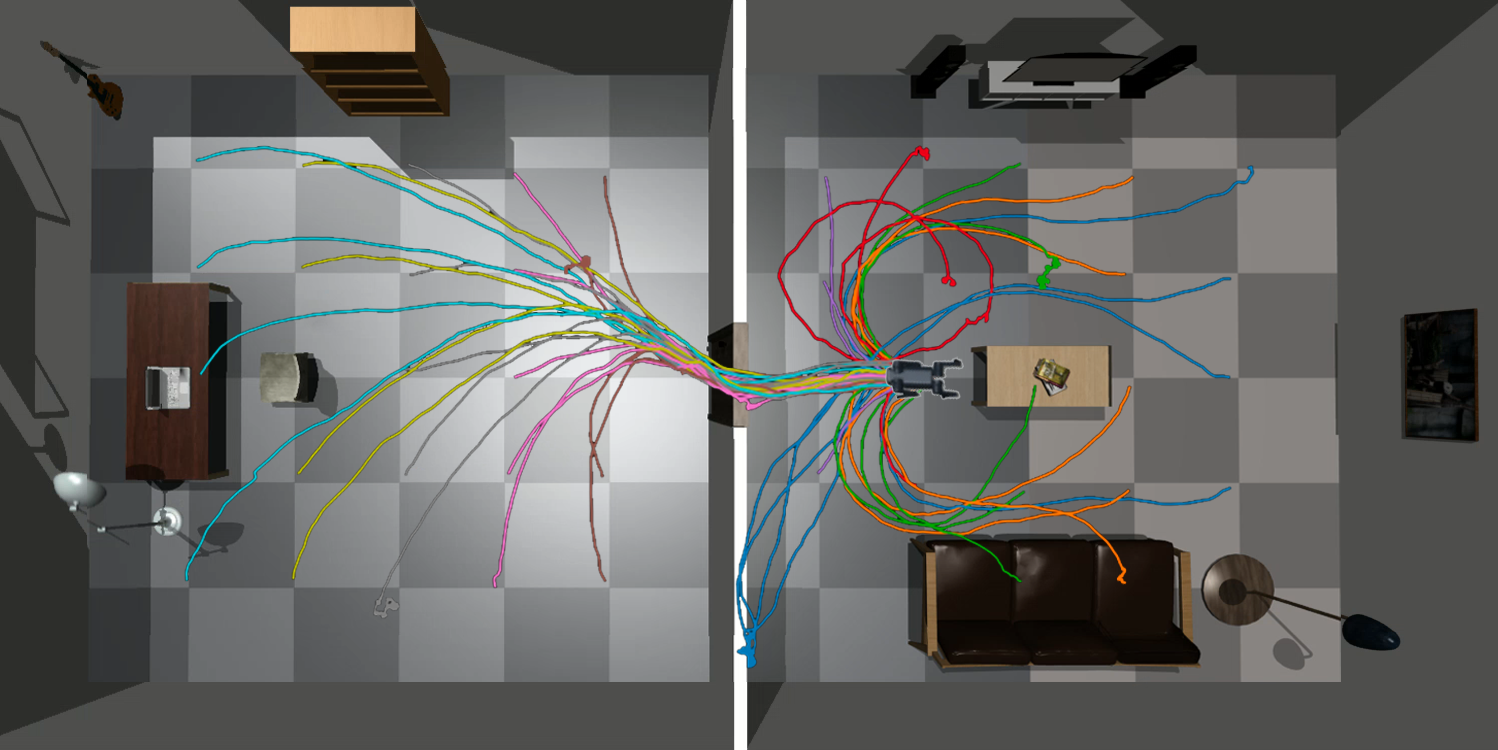}
    \caption{Trajectories taken by the robot from a fixed starting point in our Interactive Reach Environment. The target positions are sampled to be at grid corners inside the room. Our robot crawls under furniture, walks around obstacles and opens a door to get from one room to another.}
    \label{fig:trajectories}
\end{figure}

    
\subsection{Simulation Analysis} \label{sec:analysis}

This section intends to analyze the robustness and sensitivity of our learning framework, CCRL.

\noindent\textbf{Sensitivity to clumsy parent skills.} We first investigate the robustness of the proposed CCRL to check if it can learn with a clumsy, poor-performing parent policy. To this end, we learn a new policy on the \textit{Reach Target (Easy)} task, but add one more expert policy that is randomly initialized and untrained. We notice that this policy performs similarly to our best policy (CCRL) in Table~\ref{tab:success_rate}, reaching a success rate of $90\%$. This showcases the robustness of CCRL, in ignoring bad parent policies. 

\noindent\textbf{Sensitivity to irrelevant parent skills.} We also check whether CCRL is sensitive to the design of the skill graph. Particularly, we investigate if CCRL can be robust to additional unnecessary skill dependencies, i.e., additional edges. We retrain a new policy to walk in a straight line with a redundant set of skills: \{\textit{Walk straight},  \textit{Turn left}, \textit{Turn right}\}. 
Our framework ignores the redundancy and walks straight achieving an average tracking error of about $16\%$, same as the parent walking skill.

\noindent \textbf{Importance of residual penalties.} We further examine the importance of residual penalties ($L^{rm}_t$,$L^{rw}_t$) by retraining the \textit{Reach Target (Easy)} policy without residual penalties ($c_3,c_4 = 0$). With this setup, the policy ends up overusing residual actions to learn behaviors that diverge too much from the base walking skill. While it still reaches the target in many cases, the quality of trajectories is poor and unsuitable sim-to-real transfer. Please refer to the supplemental video for comparison.




\subsection{Sim2real transfer}
\begin{figure}[t]
\vspace{0.1in}
\centering
    \begin{subfigure}[h]{0.49\linewidth}
    \includegraphics[width=0.99\textwidth]{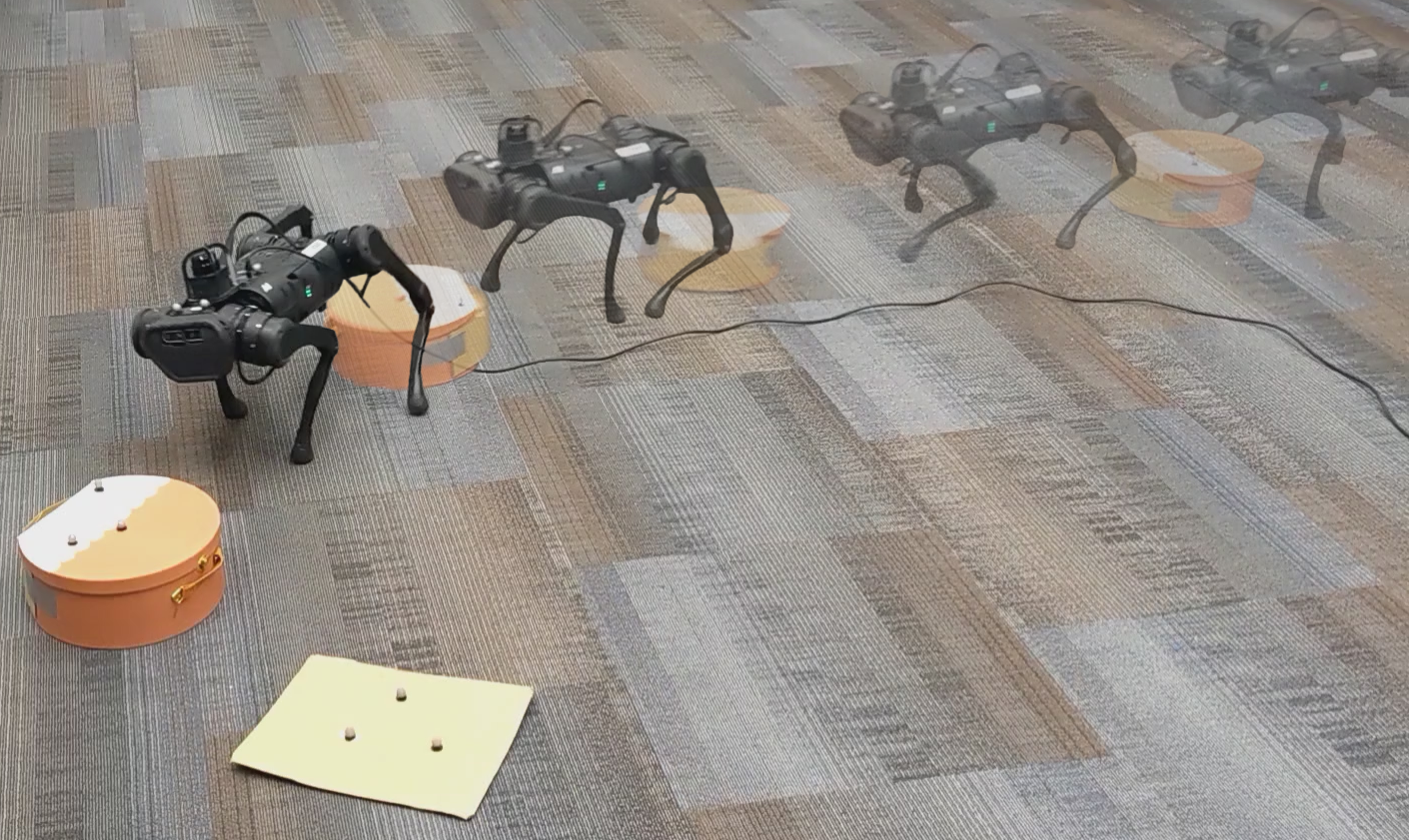}
    \caption{Push object}
    \label{fig:push_obj_real}
    \end{subfigure}
    \begin{subfigure}[h]{0.49\linewidth}
    \includegraphics[width=0.99\textwidth]{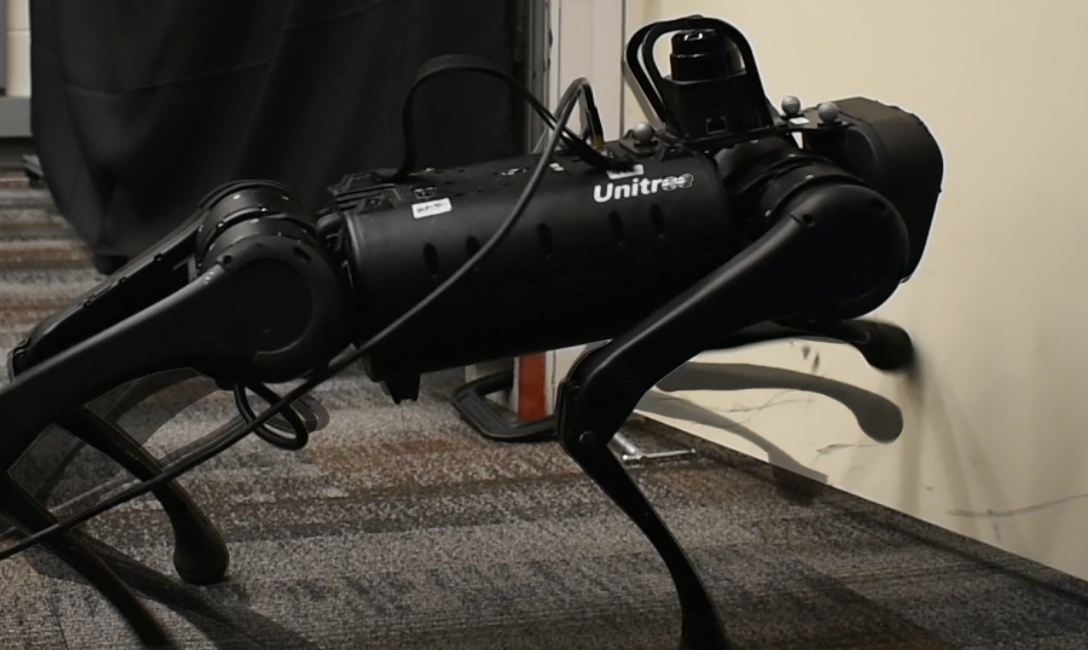}
    \caption{Open door}
    \label{fig:open_door_real}
    \end{subfigure}
    \begin{subfigure}[h]{0.98\linewidth}
    \includegraphics[width=0.99\textwidth]{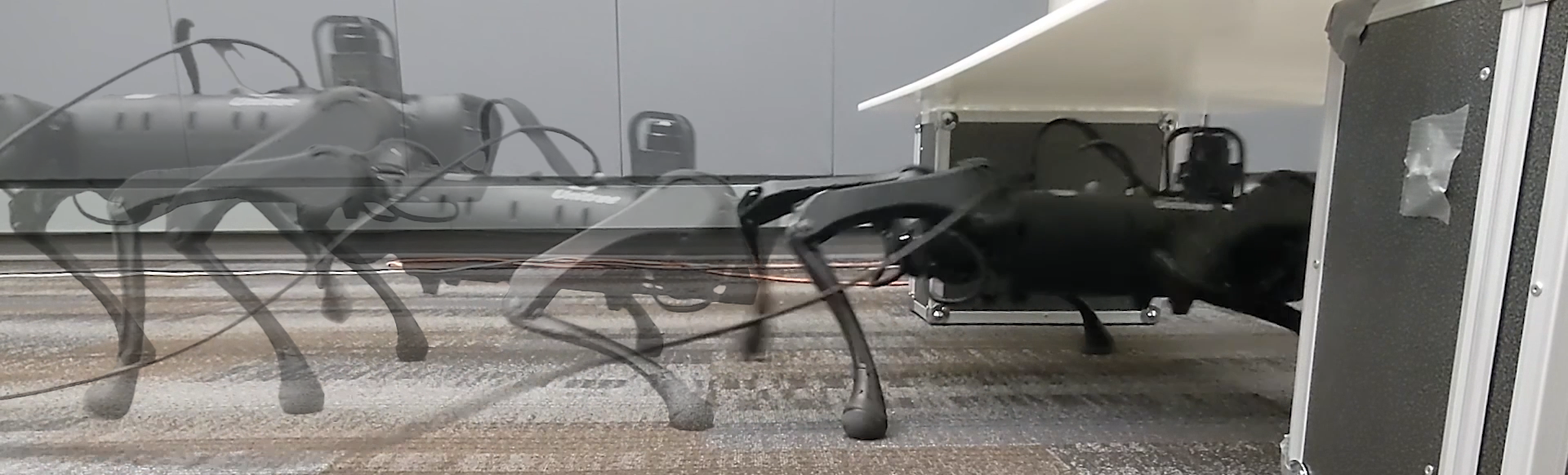}
    \caption{Crawling}
    \label{fig:open_door_real}
    \end{subfigure}
    \caption{We transfer the skills, (a) \textit{Push object}, (b) \textit{Open door}  and (c) \textit{Crawling} to the real world without any finetuning.}
    \label{fig:real_skills}
\end{figure}

We transfer our policies trained in simulation to a real Unitree A1 quadruped robot. We use motion capture to identify and track the state of the robot and objects in the real world. Our policies do not require any additional finetuning on a real robot, demonstrating the robustness of the skills generated using CCRL. In Figure~\ref{fig:real_a1_door_open} we show an interactive navigation task, where the robot crawls under a desk, reaches a door, pushes it open with its leg, and walks through it to reach a target. We show example motions of additional skills such as \emph{Push Object}, \emph{Open Door (Easy)} and \emph{Crawling} in Figure~\ref{fig:real_skills}. Please refer to the accompanied video for example motions of all the behaviors trained using our approach.
\section{CONCLUSION}
In this paper, we presented a novel Casacded Compositional Residual Learning (CCRL) framework that recursively learns policies by grounding them to a set of prerequisite skills learned in previous iterations. Using CCRL we built interactive motor controllers on a high-dimensional quadrupedal robot while ensuring that the learned policies follow a style grounded in the parent skill library. We compared the policies learned to multiple baselines and showed the effectiveness of the proposed framework. We also demonstrated sim-to-real transfer of the learned motor skills. 

For future work, we aim to automatically discover skill relationships instead of manually designing the skill graph. Additionally, we hope to extend our approach to vision-based obstacle avoidance tasks by learning image-conditioned residual perturbations to pre-trained navigation skills. 
Additionally, we want to explore interaction policies that learn about objects through interaction~\cite{xu2019densephysnet,kumar2019estimating} and adapt to the variability seen in the real world.




\bibliographystyle{IEEEtran}
\bibliography{IEEEabrv,main}
\end{document}